\documentclass[sigconf]{acmart}
\hypersetup{draft}
\usepackage{booktabs} 
\usepackage{xspace}
\usepackage{balance}
\setcopyright{rightsretained}
\usepackage[british]{babel}

\usepackage[bf]{subfigure}
\newcommand{\systemname}{{\it Rosetta}}

\makeatletter
\DeclareRobustCommand\onedot{\futurelet\@let@token\@onedot}
\def\@onedot{\ifx\@let@token.\else.\null\fi\xspace}

\def\eg{\emph{e.g}\onedot} 
\def\ie{\emph{i.e}\onedot} 
\def\cf{\emph{cf}\onedot}

\def\etal{\emph{et al}\onedot}
\makeatother

\usepackage{algorithm}
\usepackage{algpseudocode}
\usepackage{algcompatible}
\usepackage{multirow}

\acmDOI{10.475/123_4}

\acmISBN{123-4567-24-567/08/06}

\acmConference[KDD'2018]{ACM KDD conference}{August 2018}{London, United Kingdom}
\acmYear{2018}
\copyrightyear{2018}

\acmArticle{4}
\acmPrice{15.00}

\begin{document}
\title{Rosetta: Large scale system for text detection \\ and recognition in images}
\titlenote{Authors of the paper have equal contribution to this work.}

\author{Fedor Borisyuk}
\affiliation{%
  \institution{Facebook Inc.}
}

\author{Albert Gordo}
\affiliation{%
  \institution{Facebook Inc.}
}

\author{Viswanath Sivakumar}
\affiliation{%
  \institution{Facebook Inc.}
}
\renewcommand{\shortauthors}{Borisyuk, Gordo, Sivakumar}
\renewcommand{\shorttitle}{Rosetta: Large scale system for text detection and recognition in images}

\begin{abstract}
In this paper we present a deployed, scalable optical character recognition (OCR) system, which we call {\systemname}, designed to process images uploaded daily at Facebook scale.
Sharing of image content has become one of the primary ways to communicate information among internet users within social networks such as Facebook and Instagram, and the understanding of such media, including its textual information, is of paramount importance to facilitate search and recommendation applications.
We present modeling techniques for efficient detection and recognition of text in images and describe {\systemname}'s system architecture.
We perform extensive evaluation of presented technologies, explain useful practical approaches to build an OCR system at scale, and provide insightful intuitions as to why and how certain components work based on the lessons learnt during the development and deployment of the system.

\end{abstract}

%
%



\maketitle

\section{Introduction}\label{sec:intro}
One of the primary ways through which people share and consume information in social networks such as Facebook and Instagram
is through visual media such as photos and videos.
In the last several years, the volume of photos being uploaded to social media platforms
has grown exponentially to the order of hundreds of millions everyday ~\cite{MillionsOfImages}, presenting technological challenges for
processing increasing volumes of visual information.
One of the challenges in image understanding is related to the retrieval of textual information from images, also called Optical Character Recognition (OCR), which represents a process of conversion of electronic images containing typed, printed, painted or scene text into machine encoded text.
Obtaining such textual information from images is important as it facilitates many different applications, \eg search and recommendation of images.

In the task of OCR we are given an image, and the OCR system should correctly extract the text overlaid or embedded in the image.
Challenges to such a task compound as a number of potential fonts, languages, lexicons, and other
language variations including special symbols, non-dictionary words, or specific information such as URLs and email ids appear in the image,
 and images tend to vary in quality with text in the wild appearing on different backgrounds. Another aspect of the problem arises from the huge volume of images in
 social networks uploaded daily that need to be processed. Due to the nature of downstream applications, we
targeted to perform OCR in realtime once the image is uploaded, and that required us to spend significant time optimizing
the components of the system to perform within reasonable latency constrains.
Therefore, our
problem can be stated as follows: \textit{to build a robust and accurate system for optical character recognition capable of processing hundreds of millions of images per day in realtime.}

\begin{figure}[!t]
\includegraphics[height=4.15cm, clip]{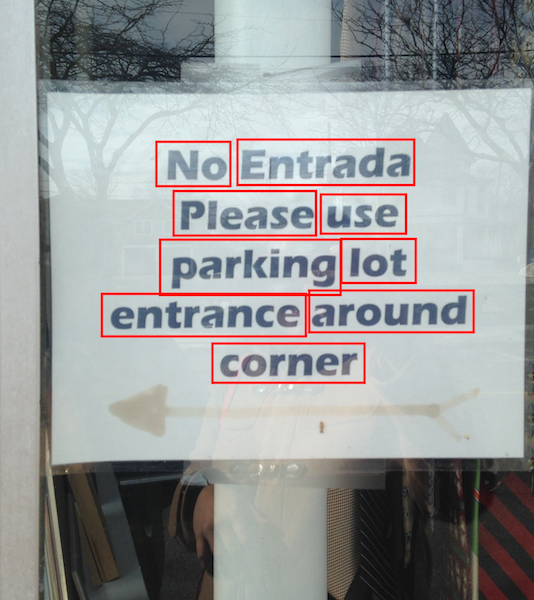}
\includegraphics[height=4.15cm, clip]{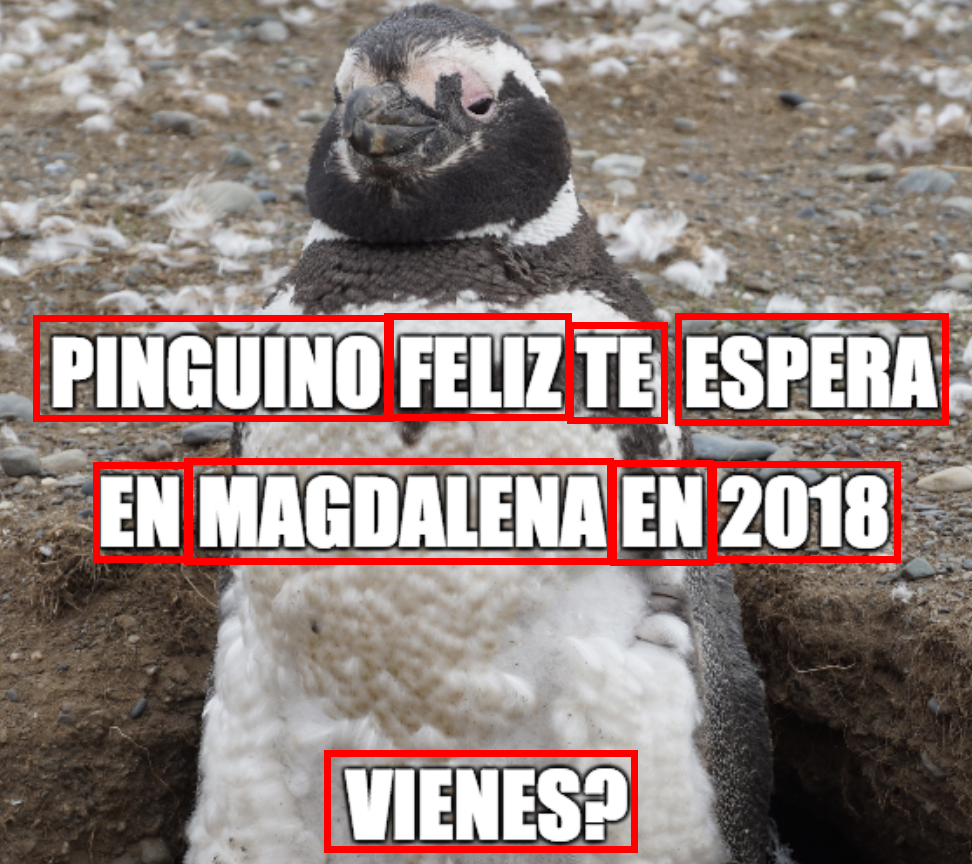}
\caption{OCR text recognition using {\systemname} system.
An approach based on Faster-RCNN detects the individual words, and
a fully-convolutional CNN produces the lexicon-free transcription of each word.}
\label{fig:sampleOCRextraction}
\vspace{-0.5em}
\end{figure}

\begin{figure*} [!t]
\centering
\includegraphics[width=0.85\textwidth]{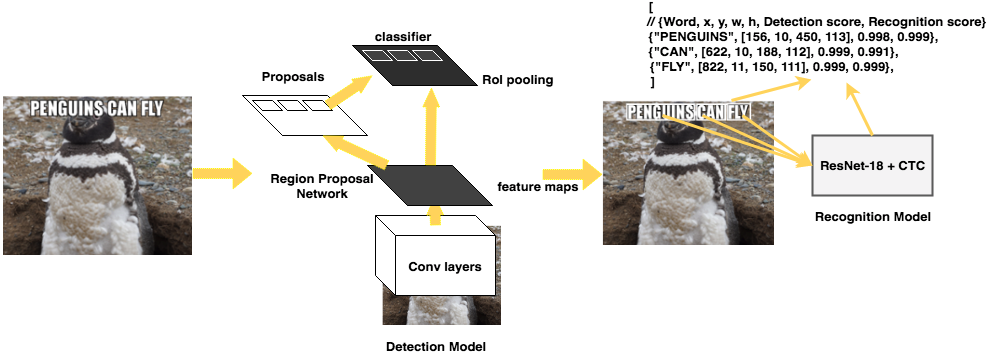}
\caption{Two-step model architecture. The first step performs word detection based on Faster-RCNN. The second step performs word recognition using a fully convolutional model with CTC loss. The two models are trained independently.}
\label{fig:model_e2e_fig}
\vspace{-0.5em}
\end{figure*}

In this paper we present {\systemname}, Facebook's scalable OCR system, which we have implemented and deployed in production and that powers downstream applications within Facebook.
Following state-of-the-art systems, our OCR is divided into a text detection stage and a text recognition stage.
Our text detection approach, based on Faster-RCNN model  ~\cite{Ren:2017:FRT:3101720.3101780}, is responsible of detecting regions of the image that contain text.
After that, our text recognition approach, that uses a fully-convolutional character-based recognition model, processes the detected locations and recognizes the text they contain.
Figure \ref{fig:sampleOCRextraction} shows some results produced by \systemname.


The rest of the paper is organized as follows:
\S\ref{sec:related_work} provides an overview of the related work.
\S\ref{sec:model} describes our detection and recognition models, and \S\ref{sec:systemarch} describes the design and architecture for \systemname,
\S\ref{sec:performancestudy} describes the experimental evaluation and the lessons learnt during the system development.
Finally, \S\ref{sec:deployment} describes the deployment and \S\ref{sec:conclusion} concludes the paper.

\section{Related Work}\label{sec:related_work}
Although the OCR problem has received a lot of attention during decades, most of this attention has been led by the document analysis community and it has been limited to document images (see \eg \cite{nagy2000pami} for some examples).
It's only been during the last decade that the computer vision community has paid more attention to the problem of text detection and recognition in natural images, mostly due to the seminal work of Wang and Belongie \cite{wang2010word}.
Since then the topic has received a significant amount of attention, and large advances in text detection and text recognition in natural images have been produced, mostly driven by the success of convolutional neural networks (CNN).

The works of Jaderberg \etal \cite{JaderbergSVZ14_synt, Jaderberg:2016:RTW:2877061.2877148} are among the first to propose a CNN based approach for text recognition to classify words
into a pre-defined set of dictionary words.
Before the recognition stage, a combination of Edge Boxes ~\cite{10.1007/978-3-319-10602-1_26}
and an aggregate channel features (ACF) detector framework ~\cite{Dollar:2014:FFP:2693345.2693405} is used to detect candidate locations for the words.
Following Jaderberg's work, many new methods have recently appeared that use CNN architectures to detect words in images, including
~\cite{DBLP:conf:cvpr:ZhouYWWZHL17, DBLP:conf/cvpr/ShiBB17, DBLP:conf/cvpr/HeYLZOKG17,  FOTSdetectionpaper}.
CNN based detection methods have the benefit of providing an end-to-end learnable framework for text detection.
We follow a convolutional approach as well to detect text in images by adapting state of the art object detection
  framework based on Faster-RCNN ~\cite{Ren:2017:FRT:3101720.3101780}.

Similarly, many recent works on text recognition have also followed Jaderberg's work \cite{Jaderberg:2016:RTW:2877061.2877148} and addressed its shortcomings.
In particularly, and contrary to the original work, most recent character-based works do not require a dictionary (also known as lexicon)
and can recognize words of arbitrary length that were not seen during training (\eg \cite{DBLP:journals/corr/ShiBY15, DBLP:journals/corr/abs-1709-01727, DBLP:journals/corr/abs-1709-04303, SqueezedText}).
In general, this is achieved by using a fully-convolutional model that given an image produces a sequence of features.
In some cases, attention mechanisms (unsupervised or supervised) or recurrent models can be leveraged to improve the quality of the features.
At training time, a sequence-to-sequence loss such as CTC (Connectionist Temporal Classification)~\cite{Graves:2006:CTC:1143844.1143891} is used to compute the error between the predicted features and the real transcription and to improve the model through backpropagation.
At testing time, the sequence of features can be converted into a sequence of characters from where the transcription is obtained, either in a greedy manner, or using language model or a dictionary.
Our fully-convolutional recognition model shares the key ideas of these works, \ie, producing a sequence of features that can be trained with CTC and that can recognize words of arbitrary length that do not appear in a dictionary.

To train such models, obtaining high quality training data is of paramount importance.
The works of ~\cite{DBLPGuptaVZ16, JaderbergSVZ14_synt, SqueezedText} provide elegant
solutions for preparation of artificial generated training dataset with hundreds to thousands of images being
annotated with printed text combining different fonts, styles and font sizes to provide a variety of data for
training robust text recognition classifiers.
We have used similar approaches in this work to generate artificial
training data for both detection and text recognition tasks.

\begin{figure*} [!h]
\centering
\includegraphics[width=1.01\textwidth]{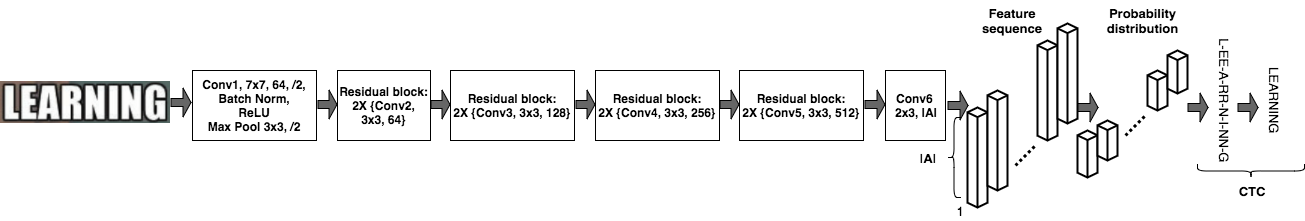}
\caption{Architecture of Text Recognition model.}
\label{fig:RecognitionModelArch}
\end{figure*}

\section{Text Extraction Models}\label{sec:model}
We perform OCR in two independent steps: detection an recognition.
In the first step we detect rectangular regions in the image potentially containing text.
In the second step we perform text recognition, where, for each of the detected regions, a CNN is used to recognize and transcribe the word in the region.
This two-step process has several benefits, including the ability to decouple training process and deployment updates to detection and recognition models, run recognition of words in parallel, and independently support text recognition for different languages.
Figure \ref{fig:model_e2e_fig} details this process.

In the following sections we describe our approach in detail.
\subsection{Text Detection Model}\label{sec:localization_model}

For text detection we adopted an approach based on Faster-RCNN ~\cite{Ren:2017:FRT:3101720.3101780}, a state-of-the-art object detection network.
In a nutshell, Faster-RCNN simultaneously performs detection and recognition by i) learning a fully-convolutional CNN that can represent an image as a convolutional feature map (typically based on ResNet architectures \cite{He2016}), ii) learning a region proposal network (RPN) that takes that feature map as input and produces a set of $k$ proposal bounding boxes that contain text with high likelihood, together with their confidence score, and iii) extracting the features from the feature map associated with the spatial extent of each candidate box, and learning a classifier to recognize them (in our case, our categories are text and no text).
To select the $k$ most promising candidates, the proposals are sorted by their confidence score, and non-maximum suppression (NMS) is used to choose the most promising proposals.
Additionally, bounding box regression is typically used to improve the accuracy of the produced bounding boxes.
The whole detection system (feature encoding, RPN, and classifiers) are trained jointly in a supervised, end-to-end manner.

Our text detection model uses Faster-RCNN, but replaces the ResNet convolutional body with a ShuffleNet-based ~\cite{DBLP:journals:corr:ZhangZLS17} architecture for efficiency reasons.
As we show empirically in Section \ref{sec:performancestudy}, ShuffleNet is significantly faster than ResNet and did not lead to inferior accuracy.
The ShuffleNet convolutional trunk is pre-trained using ImageNet dataset ~\cite{Deng09imagenet:a}.
To train the end-to-end detection system, the model was bootstrapped with an in-house synthetic dataset and then fine-tuned with COCO-Text and human annotated datasets to learn real-world characteristics, \cf  \S\ref{sec:datasec}.

\subsection{Text Recognition Model}\label{sec:recognition_model}

\begin{algorithm}[t]
\caption{Curriculum Learning for Text Recognition Model}
\label{alg:curriculumtraining}
\begin{flushleft}
\textbf{Input:} Training dataset $D$; Number of warm-up epochs $W$; Number of epochs $N$; Initial maximum length of words $l_0$; Warm up word width $w_w$; Initial word width $w_0$; Learning rates $\alpha$ and $\beta$; Learning rate decay period $t$\\
\textbf{Output:} Trained CTC model. \hspace{1in}\\
\end{flushleft}
\begin{algorithmic}[1]
\STATE \emph{Note:} The values of learning rates $\alpha$ and $\beta$ are determined empirically by observing training convergence.
$\alpha$ is the highest learning rate that allows one to train the model after random initialization without diverging,
 while $\beta$ is the highest learning rate that allows one to train the model after initializing with pretrained weights.
\STATE Set initial learning rate $lr=\alpha$.
\STATE Set log-increment $\Delta = \frac{\log{\beta} - \log{\alpha}}{N}$.
\STATE Set image width preprocessing to $w_w$
\STATE Warmup training:
\STATE Set $l=l_0$.
\FOR{($i=1$; $i \leq W$; $i++$, $l++$)}
\STATE Generate filtered training dataset $T$ by keeping only training examples from $D$ with labeled
word length equal or smaller than $l$ characters.
\STATE Train CTC model for one epoch using filtered training dataset $T$ and learning rate $lr$.
\STATE Increase learning rate $lr = \alpha + 10^{i\Delta}$
\ENDFOR
\STATE Post-warmup training:
\STATE Set $w=w_0$
\FOR{($i=1$; $i \leq N$; $i++$, $w+=8$)}
\STATE Set image width preprocessing to $w$. Generate filtered training dataset $T$ by keeping only training examples from $D$ with labeled
word length equal or smaller than the size of the feature map when using width $w$.
\STATE Train CTC model for one epoch using filtered training dataset $T$ and learning rate $lr$.
\STATE Set $lr = lr * 10^{-floor(i / t)}$
\ENDFOR
\STATE Return model.
\end{algorithmic}
\vspace{-0.5ex}
\end{algorithm}

We experimented with different architectures and losses for text recognition.
The first one is based on the character sequence encoding model (CHAR) of Jaderberg \etal.~\cite{JaderbergSVZ14_synt}.
The model assumes that all images have the same size (resized to 32x100 in
~\cite{JaderbergSVZ14_synt} without preserving the aspect ratio) and that there is a
  maximum number of characters per word $k$ that can be recognized ($k=23$ in ~\cite{JaderbergSVZ14_synt}).
For longer words, only $k$ of its characters will be recognized.
The body of the CHAR model consists of a series of convolutions followed by $k$ independent multiclass classification heads,
each of which predicts the character of the alphabet (including the NULL character) at each position.
During training, one jointly learns the convolutional body and the $k$ different classifiers.

The CHAR model is easy to train using $k$ parallel losses (softmax + negative cross-entropy) and provides a reasonable baseline,
but has two important drawbacks: it can't recognize correctly words that are too long (for example URLs),
and the number of parameters in the classifiers is very large, leading to big models that tend to overfit.

\begin{figure*}[!h]
\centering
\includegraphics[width=\linewidth, clip]{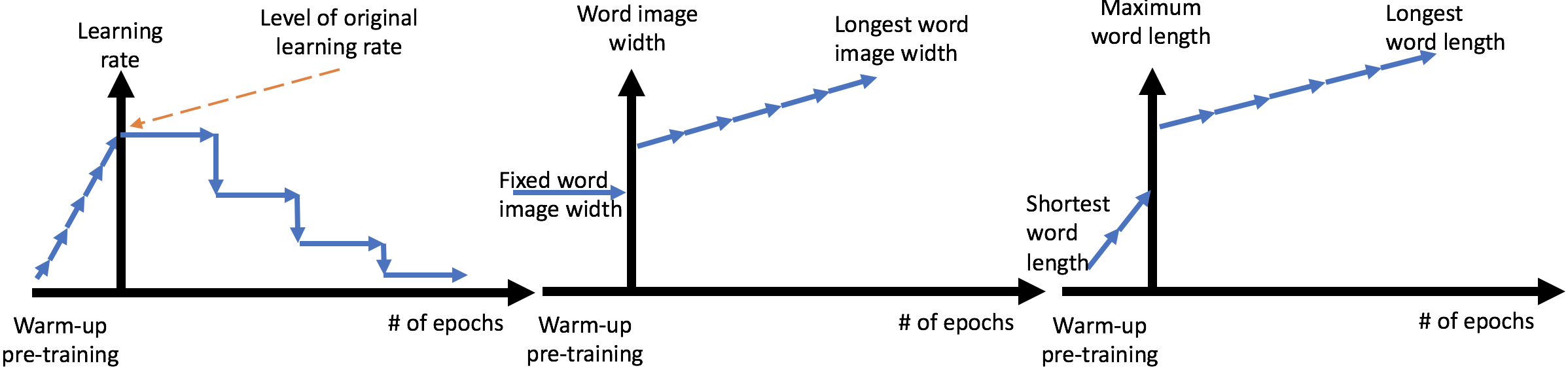}
\caption{Schematic visualization for the behavior of learning rate, image width, and maximum word length under curriculum learning for CTC text recognition model.}
\label{fig:CurriculumLearning}
\end{figure*}

The second model architecture is a fully-convolutional model that outputs a sequence of characters.
We refer to this model as CTC because it uses a sequence-to-sequence CTC loss~\cite{Graves:2006:CTC:1143844.1143891} during training.
The general architecture of the CTC model is illustrated in Figure ~\ref{fig:RecognitionModelArch}.
After the convolutional body, which is based on a ResNet-18 ~\cite{He2016} architecture, it has a last convolutional layer that predicts the most likely character at every image position of the input word.
Different from other works, we do not use an explicit recurrent network (such as an LSTM or GRU) stage or any attention mechanism, and produce directly
the probability of each character.
As noted by recent works (\eg \cite{DBLP:journals/corr/abs-1709-04303}), convolutions can still model the interaction between characters while being computationally more efficient at test time.

 This model is trained using the CTC loss, which computes the conditional probability of a label given the prediction by marginalizing over the set of all possible alignments paths,
  and that can be efficiently computed using dynamic programming.
  As shown in Figure ~\ref{fig:RecognitionModelArch}, every column of the feature map corresponds to the probability distribution of all characters of the alphabet at that position of the image,
and CTC finds the alignments between those predictions, which may contain duplicate characters or a blank character ($-$), and the ground truth label.
For example, in Figure ~\ref{fig:RecognitionModelArch}
we show that for the input training word $LEARNING$, the model might produce the sequence of characters "L-EE-A-RR-N-I-NN-G",
which includes blanks and duplicates.

At inference time, computing the optimal labeling is typically intractable.
Instead, one typically relaxes the problem:
 based on the assumption that the most probable path will correspond
to the most probable labelling, one can find the best path decoding in linear time by greedily taking the most likely character at every position of the sequence.
As a postprocessing, one then removes contiguous duplicate characters not delimited by the blank character.

This CTC model addresses the two problems present in the CHAR model, \ie, it has significantly less parameters (because it does not require $k$ independent fully-connected layers) and can predict words of arbitrary length (because the model is fully-convolutional).
We also compared inference time of CHAR model versus CTC model and observed that the CTC model is 26\% faster than the CHAR model, mostly due to the cost of the $k$ fully connected layers that perform character classification in the CHAR model.
Orthogonal to this, one can use different body architectures.
We experimented with ResNet-18 and SqueezeNet
 ~\cite{DBLP:journals/corr/IandolaMAHDK16}, and obtained higher accuracy with ResNet-18 while experiencing only a very small overhead in computation.

\textbf{Image Pre-processing:} In most approaches to text recognition (starting with Jaderberg \etal~\cite{JaderbergSVZ14_synt}), the word image crops are preprocessed by
resizing them to 32x100 pixels without preserving the aspect ratio.
 This leads to same characters looking very different depending on the length of the word, which in turn requires more model capacity to accurately represent.
Instead, at training time, we resize the word images to 32x128 pixels distorting the aspect ratio only if they are wider that
  128 pixels, and using right-zero-padding otherwise. This ensures that, for most words, no distortion is produced.
  Setting a fixed width is necessary from a practical point of view to be able to efficiently train using batches of images.

  At testing time we resize the images to a height of 32 pixels preserving their aspect ratio (and so they can have an arbitrarily long width).
  This ensures that test and train images don't have, in general, a large domain shift.
  In preliminary experiments we noticed this approach to yield superior accuracy than to resize the images without preserving the aspect ration.

Additionally, it is interesting to note that given a body of the neural network (in our case ResNet-18) and
a word image, the number of character probabilities emitted depends on the width of the word image, not on the number of characters of the word (although obviously both are correlated).
By stretching all training and testing images by the same, constant factor, we can control the number of probabilities emitted and approximately align it with the length of the transcription, which can affect training convergence and testing accuracy.
We empirically found that a stretching factor of 1.2 leads to superior results than using the original aspect ratio.

\textbf{Training of Text Recognition Model:} We firstly train our text recognition models using synthetic data and
 then fine-tune on the application specific human rated data (described in \S\ref{sec:datasec}) to achieve better results.

Interestingly, we found the CTC model much harder to train than the CHAR model: while the CHAR model converged quite rapidly and was not very sensitive to initial parameters such as the learning rate,
the CTC model consistently either diverged after just a few iterations or trained too slow to be of practical use.
We speculate that aligning sequences is a much harder problem than learning independent classifiers, and that that's the reason for the model not to train satisfactorily.

To effectively train the CTC model we considered two workarounds. The first one was to initialize the weights of the model body with the trained weights of the CHAR model, and then finetune those weights while simultaneously learning the last convolutional layer from scratch.
The second approach was based on curriculum learning~\cite{Bengio2009}, \ie, starting with a simpler problem and increasing the difficulty as the model improves.
 Instead of training with all training words directly, we warmed up the model for 10 tiny epochs that considered only very short words.
We started training with words of length $\leq 3$, for which the alignment would be easy and where the variations in length would be tiny, and increased the maximum length of the word at every epoch.
We also reduced the width of the images to simplify the initial problem.
Simultaneously, we started with a tiny learning rate and kept increasing it at every epoch until it reached the same initial learning rate of the CHAR model.
This idea of increasing the learning rate during warm up has been used in other cases, see \eg \cite{goyal2017hour}.
After this warmup standard training would follow, decreasing the learning rate after $n$ epochs.
However, even at this stage, we still apply the principles of curriculum learning:
at the end of every epoch, besides the reduction of learning rate, we also increase the width of the words
\footnote{And, as a byproduct, the maximum length of the words that can be used for training, as the maximum accepted length is given by the size of the feature map, which in turn depends on the width of the image words.}.
A representation of the changes in learning rate and word length can be seen in Figure \ref{fig:CurriculumLearning}, while Algorithm ~\ref{alg:curriculumtraining} describes the algorithm in detail.

Both the pretraining and the curriculum learning approaches worked well and led to very similar quantitative results.
 In the end, we adopted curriculum learning as our standard approach because of two reasons: first, it was slightly faster to perform the warm up than to train the CHAR model from start to end.
Second, and most important, curriculum learning helped us to eliminate the need of training
a separate CHAR model and copying the weights to the CTC model, which created an undesirable dependency between the models.

\begin{figure*} [!h]
\centering
\includegraphics[width=0.7\textwidth]{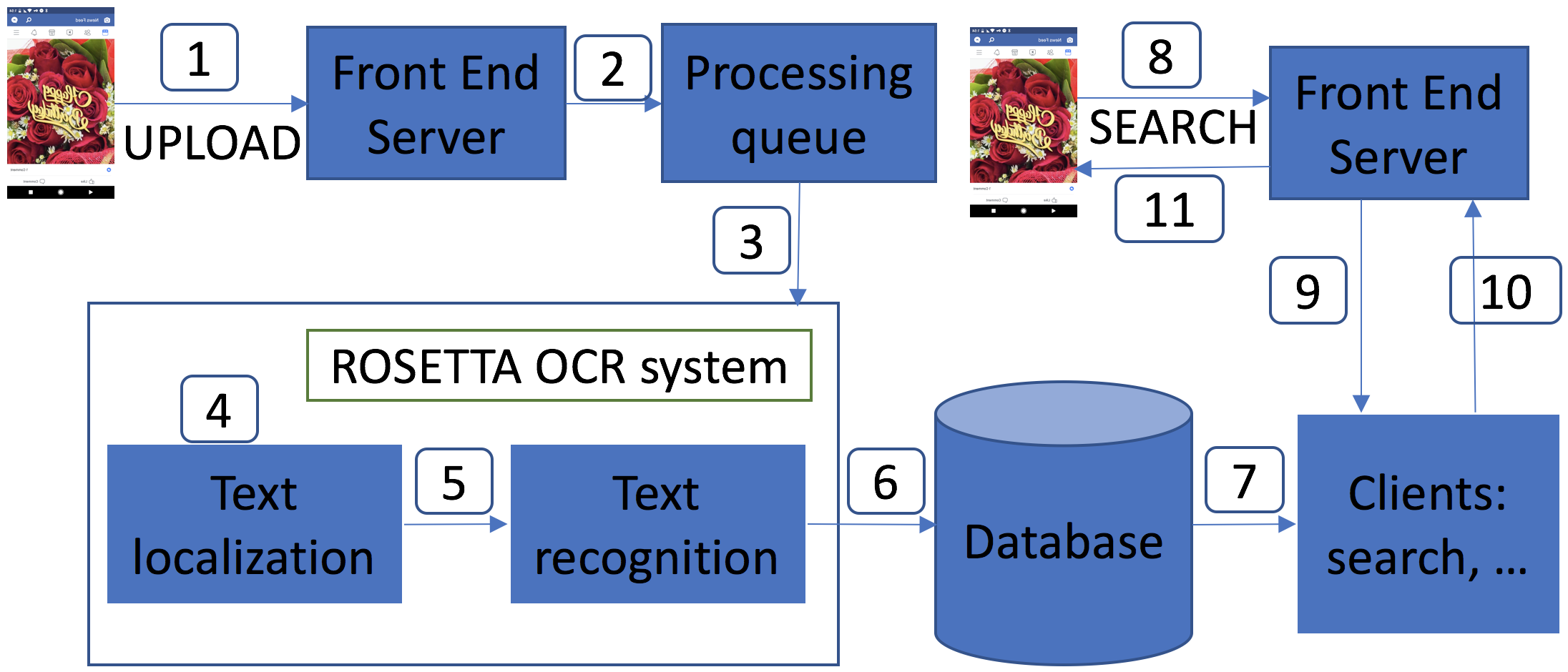}
\caption{Architecture of {\systemname}, Facebook's scalable text recognition system.}
\label{fig:ArchitectureFig}
\end{figure*}

\section{System Architecture}\label{sec:systemarch}
We now describe the system architecture of \systemname, Facebook's realtime large-scale OCR system.
{\systemname} is deployed in production and is designed to operate on images uploaded daily at Facebook scale in a realtime fashion.
Figure \ref{fig:ArchitectureFig} outlines the high-level architecture of {\systemname}.
 {\systemname} utilizes a pull-based model where an image uploaded by a client application (step 1 in Figure \ref{fig:ArchitectureFig}) is added to a distributed processing queue.
The inference machines in {\systemname} pull the enqueued jobs when resources are available and process them asynchronously.
Consumers can register callbacks when enqueuing jobs, which {\systemname} invokes right after each job is processed for
immediate usage of results by downstream applications.
The processing queue is optimized for high throughput and availability, and utilizes RocksDB ~\cite{rocksdb} underneath for persistence.
This pull-based asynchronous architecture provides various benefits including better load-balancing, rate-limiting in scenarios of unexpected spikes in requests (for example, surge in photo uploads on Facebook), and the ability to optimize the system for throughput.

The online image processing within {\systemname} consists of the following steps:
\vspace{-0.2em} 
\begin{enumerate}
\itemsep0em
\item The image is downloaded to a local machine within {\systemname} cluster and pre-processing steps such as resizing and normalization are performed.
\item The text detection model is executed (step 4 in Figure \ref{fig:ArchitectureFig}) to obtain location information (bounding box coordinates and scores) for all the words in the image.
\item The word location information is then passed to the text recognition model (step 5 in Figure \ref{fig:ArchitectureFig}) that extracts characters given each cropped word region from the image.
\item The extracted textual information along with the location of the text in the image is stored in TAO, Facebook's distributed graph database ~\cite{Bronson:2013:TFD:2535461.2535468} (step 6 in Figure \ref{fig:ArchitectureFig}).
\item Downstream applications such as Search can access the extracted textual information corresponding to the image directly from TAO (step 7 in Figure \ref{fig:ArchitectureFig}).
\end{enumerate}

In section \S\ref{sec:performancestudy}, we show various experiments performed that helped us make the right trade-offs between accuracy of the system and inference speed during the development of {\systemname}.

\section{Experiments}\label{sec:performancestudy}
We present an extensive evaluation of {\systemname} OCR system. Firstly we define the metrics used to judge the accuracy and processing times of the system, and describe the datasets used for training and evaluation. We follow the standard practice of training and evaluating our models on separate holdout datasets.
We describe the evaluation of the proposed text detection and text recognition models, and explain the design decisions made for accuracy vs inference speed trade-offs.
 Finally we share various lessons learnt during the development and deployment of {\systemname} that helped us launch at Facebook scale.




\subsection{Metrics}\label{sec:metricssec}
We used a combination of the following accuracy and inference time metrics to evaluate {\systemname}.
We report all metrics relative to the baseline and some of the measurements are reported normalized to the interval $[0,1]$ by min-max normalization.
Most of the reported experiments are presented in the form of ablation study, where we list improvements in metrics over the baseline model configuration or baseline variant of the training data.

\textbf{Performance of detection model}: For detection tasks, the standard metric to measure performance is Mean Average Precision (mAP) at certain IoU (Intersection-over-Union) thresholds.
The IoU between a predicted bounding box region and a ground-truth region is defined as the ratio of the area of the intersection of the boxes to the area of the union of the boxes.
 While precision and recall are single-valued metrics that measure performance at a given threshold, Average Precision (AP) takes into account the order of the predictions as well.
Average Precision (AP) measures the area under the precision-recall curve for a given IoU threshold interval.
Formulaically, $AP = \sum_{n} (R_{n} - R_{n-1})P_{n}$, where $P_{n}$ and $R_{n}$ are the precision and recall at the $n^{th}$ score threshold.
Mean Average Precision (mAP) is calculated as average of AP's across the whole test set.
As is standard for detection tasks, we report mAP with a 50\% IoU threshold (mAP@0.5) and mAP@0.5:0.95 as the evaluation metric for quality of text detection model.
mAP@0.5:0.95 is calculated as average of set of mAP's over different IoU thresholds, from 0.5 to 0.95 with step 0.05 (0.5, 0.55, 0.6 ..., 0.95).
Higher mAP value for the detection model is desirable.

\textbf{Performance of text recognition model}: For the recognition model, we use accuracy and Levenshtein's edit distance as metrics to measure the performance.
 The accuracy is calculated based on statistics of prediction across test set by $Accuracy = \frac{TP+TN}{TP+FP+FN+TN}$, where TP, TN, FP, and FN are the true positives, true negatives, false positives, and false negatives.
For accuracy calculation, a prediction is counted as positive only if the predicted word matches exactly the ground-truth word.
The accuracy metric gives information about the percentage of words that are correctly recognized, but does not give any information about how wrong the incorrect predictions are: incorrectly recognizing one character or incorrectly recognizing ten characters yields exactly the same accuracy, zero.
To obtain more information about the incorrect transcriptions we use Levenshtein's edit distance, which counts the number of single character edits (insertions, deletions and substitutions) between the predicted word and the ground-truth word, and can be seen as a surrogate of how much work would be needed to manually correct the transcription.
We report the total edit distance for all items in the test set.
Higher accuracy and lower edit distance are desirable.

\textbf{Inference time}: We used internal Facebook infrastructure to measure the inference runtimes of detection and recognition models.
Inference time for detection model represents the time spent detecting and calculating the bounding boxes for all textual information observed in the image.
Inference time for recognition model represents the time spent running the recognition model on every detected word in an image.
The experiments were performed on hardware settings closely matching the production setup and using only one cpu and one single core.
Lower inference times are desirable.

\subsection{Training and Test Datasets}\label{sec:datasec}

Having high quality training and test data is important for building robust supervised machine learning models.
We used the COCO-Text ~\cite{COCOText} public dataset, which contains extensive annotations of text in the wild, to bootstrap training.
COCO-Text contains more than 63,000 images and 145,000 text instances.
However COCO-Text doesn't match the data-distribution of images uploaded to Facebook.
For instance, a non-trivial portion of images on Facebook have text overlaid on them which COCO-Text fails to adequately capture.
To address this, we generated a large synthetic dataset taking into account different use-cases.
This was done by overlaying words on randomly selected public Facebook images.
The images where filtered to select only those that do not already have text on them by using a separate image classification CNN that predicts the probability of text being present in an image.
Image generation is part of the training process, where we automatically prepare the set of images at training initialization step, and remove the images after completion of the model training.
The words to generate this synthetic dataset were picked at random from a dictionary, and were augmented to include special characters and to look like email addresses, URLs and phone numbers,
 and then overlaid on images with various fonts, sizes and distortions applied.
We used around 400k synthetic images for training and 50k images for testing.
We also manually annotated a dataset with thousands of images gathered using help of human raters and used it to fine-tune the models, which greatly improved the results.
In the following sections of experimental evaluation we report most of the metrics on the human rated dataset mentioned here.

\begin{table}[!h]
\centering
\caption{mAP of Faster-RCNN detection model on the test set with different training datasets.
 Accuracies reported as relative improvements of mAP over training only on synthetic dataset.
 The $\rightarrow$ denotes finetuning, \ie, $A \rightarrow B$ means train on A and then finetune on B.}
\label{tab:tablemetrics_detection_data}
\begin{tabular}{lll}
\toprule
& \multicolumn{2}{c}{mAP (relative improvement)} \\
\cmidrule{2-3}
Training dataset & @IOU=0.5 & @IOU=0.5:0.95 \\
\midrule
Synthetic & +0.0\% & +0.0\% \\
COCO-Text &  +39.9\% & +15.2\%\\
Synthetic $\rightarrow$ COCO-Text &  +41.2\% & +16.6\% \\
Synthetic $\rightarrow$ COCO-Text  & \multirow{2}{*}{+57.1\%} & \multirow{2}{*}{+35.2\%} \\
$\quad \rightarrow$ Human rated & & \\
\bottomrule
\end{tabular}
\vspace{-1.0em}
\end{table}

\begin{table}
\centering
\caption{Inference runtimes of {Faster-RCNN} with various convolutional bodies. Numbers in the table reported as relative improvements to ResNet-50.}
\label{tab:tableLocalizationNetBodySpeed}
\begin{tabular}{lll}
\toprule
Convolutional body & Ratio of CPU Inference time \\
\midrule
ResNet-50 & 1x \\
ResNet-18 & 1.9x faster \\
ShuffleNet & 4.57x faster \\
\bottomrule
\end{tabular}
\vspace{-1.0em}
\end{table}

\begin{table}[!h]
\centering
\caption{mAP of {Faster-RCNN} with ResNet-18 and ShuffleNet bodies evaluated on COCO-Text dataset.
 mAP numbers in the table reported as relative improvements to ResNet-18 with 3 RPN aspect ratios.}
\label{tab:tableLocalizationNetBody}
\begin{tabular}{lllll}
\bottomrule
& & \multicolumn{2}{c}{mAP (relative improvement)} & \\
\cmidrule{3-4}
Convolutional &RPN Aspect& @IOU=0.5 & @IOU=0.5:0.95 \\
body & Ratios & & \\
\midrule
ResNet-18 & 3 & +0.0\% & +0.0\%  \\
ResNet-18 & 5 & -3.4\% & -1.1\%  \\
ResNet-18 & 7 & +2.4\% & +1.4\% \\
\midrule
ShuffleNet & 3 & +0.7\% & +0.6\% \\
ShuffleNet & 5 & +0.3\% & +0.4\% \\
ShuffleNet & 7 & +3.1\% & +1.8\% \\
\bottomrule
\end{tabular}
\vspace{-1.0em}
\end{table}

\begin{table}[!h]
\centering
\caption{mAP of Faster-RCNN detection model on test set varying RPN\_POST\_NMS\_TOP\_N (number of top RPN proposal boxes to retain). Inference is 2x faster with 100 proposals compared to 1000. Numbers in the table reported as relative improvements to configuration of Faster-RCNN with RPN\_POST\_NMS\_TOP\_N=50.}
\label{tab:paramAnalysisDetection}
\begin{tabular}{lll}
\toprule
& \multicolumn{2}{c}{mAP (relative improvement)} \\
\cmidrule{2-3}
RPN Post NMS Top N & @IOU=0.5 & @IOU=0.5:0.95 \\
\midrule
50 & +0.0\% & +0.0\% \\
100 &  +5.9\% & +2.3\% \\
1000 &  +8.2\%& +2.7\% \\
\bottomrule
\end{tabular}
\vspace{-1.0em}
\end{table}

\begin{table}
\centering
\caption{mAP of Faster-RCNN detection model on test set depending on different NMS methods for suppressing bounding boxes produced by the final regression head.
An Intersection-over-Union (IoU) threshold of 0.7 is used across all settings. For Gaussian SoftNMS, we set $\sigma$=0.5 as in ~\cite{DBLP:journals:corr:BodlaSCD17}. Numbers in the table reported as relative improvements to Standard NMS configuration.}
\label{tab:tableSOFTNMS}
\begin{tabular}{lll}
\toprule
& \multicolumn{2}{c}{mAP (relative improvement)} \\
\cmidrule{2-3}
NMS Method& @IOU=0.5 & @IOU=0.5:0.95 \\
\midrule
Standard NMS & +0.0\% & +0.0\% \\
Gaussian SoftNMS & +1.3\% & +0.9\% \\
Linear SoftNMS &   +1.5\%& +1.0\% \\
\bottomrule
\end{tabular}
\vspace{-1.0em}
\end{table}

\begin{table*}[!h]
\centering
\caption{Recognition model performance on different datasets. Higher accuracy and lower edit distance are better. Values in edit distance column are normalized by maximum value of occurred error for CHAR+Synthetic variant. Numbers in the table reported as relative improvements to configuration of CHAR model trained on Synthetic dataset.}
\label{tab:tablerecognitionmetrics}
\begin{tabular}{llll}
\toprule
& & & Normalized  Relative Reduction \\
Model& Training & Relative Accuracy & of Edit Distance\\
\midrule
CHAR & Synthetic & +0.0\% &   -0.0\% \\
CTC & Synthetic & +6.76\% & -11.23\%\\
CHAR & Synthetic $\rightarrow$ Human rated & +42.19\%& -67.01\% \\
CTC & Synthetic  $\rightarrow$ Human rated &  +48.06\%& -78.17\% \\
\bottomrule
\end{tabular}
\vspace{-1.0em}
\end{table*}

\begin{table}
\centering
\caption{Normalized magnitude of drop of recall of detected words for combination of detection and recognition systems. Normalization is performed by taking a relative drop in recall and normalizing it by the maximum value of drop for "Case sensitive". The $^*$ symbol denotes that recognitions with one character error are still considered correct.}
\label{tab:end2endmetrics}
\begin{tabular}{lll}
\toprule
 & Normalized magnitude of recall drop \\
 & in word recognition \\
 \midrule
Case sensitive & -1X \\  
Case insensitive & -0.94X \\  
Case insensitive$^*$& -0.63X \\  
\bottomrule
\end{tabular}
\vspace{-1.0em}
\end{table}

\subsection{Detection Model Experiments}\label{sec:detection_exp}
Text detection part of the implemented system is the most compute and latency intensive component during inference.
Given our scale and throughput requirements, we spent significant amount of time improving the execution speed of text detection model while keeping the detection accuracy high.
As we evaluated various approaches for the detection model, Faster-RCNN was a natural choice owing to its state-of-the-art results and readily available implementation within Facebook through
Detectron~\cite{DetectronOSS}. Detectron is Facebook's state-of-the-art platform for object detection research.
Detectron is open-source~\cite{DetectronOSS} and we refer to its settings in this section while describing our experiments.

We evaluated various architectures for the convolutional body of Faster-RCNN including ResNet-50, ResNet-18 and
ShuffleNet. Text detection model based on Faster-RCNN with ResNet-50 body incurs significant and impractical runtime on CPU per
image whereas ShuffleNet is 4.5x faster. We show comparison of inference times for different convolutional bodies for Text Detection in Table \ref{tab:tableLocalizationNetBodySpeed}.
We further evaluated mAP of ResNet-18 and ShuffleNet on COCO-Text dataset as listed in Table \ref{tab:tableLocalizationNetBody}, where we report mAP metrics as relative improvements to ResNet-18 with 3 RPN aspect ratios. ShuffleNet achieves competitive results in mAP in comparison to ResNet-18.
The region proposal network (RPN) of Faster-RCNN was tweaked to
generate wider proposals to handle text boundaries by modifying RPN.ASPECT\_RATIOS setting in Detectron
(with 7 aspect ratios and 5 different sizes, the RPN generates 35 anchor boxes per region), which showed consistently better results than the standard aspect ratios of 0.5, 1 and 2.
All further experiments were performed on ShuffleNet since optimal inference latency was one of our primary goals.

We started development of the text detection model using COCO-Text dataset. However, we quickly observed that many applications of {\systemname} have different production data distributions,
which posed challenges during model development. For example, some applications have many images with printed text overlaid on top of original photos,
where models trained just on COCO-Text showed suboptimal performance. At the same time, there were other applications where images with text occurring in natural circumstances is the norm.
This was an \textbf{important lesson} we learnt in the beginning of our development, so we decided to introduce several datasets to solve the problem. Our solution to the problem was
that we introduced artificially generated dataset with text overlaid images, and pre-trained models with it initially.
Both detection and text recognition models were pre-trained on artificial dataset and then fine-tuned using COCO-text and
human rated datasets specifically collected for client applications. In Table \ref{tab:tablemetrics_detection_data}, we present an evaluation of text detection model
improvements with fine-tuning. We initially trained using synthetic dataset, then fine-tuned using COCO-Text, and finally with human rated dataset achieving 57\% improvement
in mAP over the model trained on the synthetic dataset alone.

During inference, RPN\_POST\_NMS\_TOP\_N setting of Detectron, which controls the number of generated proposals to be fed through RoI Pooling and the final stage of detection network,
was reduced to 100 from the default 1000 as shown in Table \ref{tab:paramAnalysisDetection}. This made inference 2x faster with acceptable drop in mAP, while reducing it
below 100 gave diminishing returns in terms of inference time.

Another trade-off between performance and inference time of the detection model is related to
the resolution of the input image.
We experimented with a range of image resolutions by resizing the image to a particular size in the maximum dimension
while maintaining its aspect ratio. We found that having images of side more than 800px increases the mAP@0.5 slightly, but significantly increases inference time.

We also experimented with a modification of NMS (Non Maximal Suppression) by replacing the final NMS in Faster-RCNN with
SoftNMS following \cite{DBLP:journals:corr:BodlaSCD17}. Result of the experiments are shown in Table \ref{tab:tableSOFTNMS},
where we report relative gains in mAP@0.5 by using different strategies of SoftNMS. Overall we improved mAP@0.5 by absolute 1.5 points using SoftNMS.

\textbf{One lesson} that we learnt during the development of text detection model is related to the choice of metric for model's performance. Our initial choice of metric for Faster-RCNN model training was F1-score (a harmonic mean of precision and recall), which was at the initial moment of development a default evaluation metric in COCO-Text \cite{COCOText}. However, we observed that our reported validation set F1-score would fluctuate with a standard deviation of 5.87 points among training runs with the same configuration. We found the reason for the observed variation in performance to be because F1-score is measured at a single point in the precision-recall curve, which means we need to pick a threshold on the curve to compute it. With minor variations in training runs such as random initialization of weights, the threshold fluctuates as well and is not a reliable way to measure the model's performance, while Average Precision on the other hand computes the area under the precision-recall curve and hence is robust to such variations. At the moment of writing this paper, newer tasks on COCO-Text such as ICDAR ~\cite{ICDAR2017} use mAP for evaluation. Therefore, we have replaced F1-score with mAP and that helped to solve the problem.

\subsection{Recognition Model Experiments}\label{sec:recongnition_exp}
During the development of {\systemname} we considered different architectural designs for the
text recognition model, ranging from dictionary-based approaches where word recognition is modeled as a classification
task amongst a predefined set of words
~\cite{Jaderberg:2016:RTW:2877061.2877148}, to character based approaches
 where all the characters are jointly recognized and the combination of those characters comprises the whole word.
 Dictionary based approaches have been shown to yield superior accuracy than character
 based word recognition when a dictionary or lexicon is known in advance,
mainly due to the capacity of models to memorize a large yet limited set of dictionary words during training.

 However, a predefined dictionary would be too limiting for many real-world applications, which require
 recognizing more than just simple words as in the case of URLs, emails, special symbols and different languages.
 Therefore, \textbf{an important architectural decision} and a natural choice was to use a character-based recognition model.

In the initial stages of development, we tested various efficient backbone architectures,
including ResNet-18 \cite{He2016} and SqueezeNet ~\cite{DBLP:journals/corr/IandolaMAHDK16}.
We evaluated these choices on our synthetic data and ended up selecting ResNet-18 as our main backbone architecture for the recognition body:
 ResNet-18 showed significantly better accuracy compared to SqueezeNet, while incurring only a negligible overhead in computation.
The discrepancy between detection and recognition models with respect to the inference performance of ResNet-18 is
because, compared to the detection model, the recognition model operates on much smaller input images: while the detection model operates on resolutions close to $600\times 800$ pixels, the recognition one operates on $32\times 128$ pixels.
While the larger images allow architectures like SqueezeNet or ShuffleNet to show how efficient they are, the overheads in recognizing smaller images significantly reduce this efficiency gap.
Therefore all the experimental numbers for text recognition model provided in this section are based on ResNet-18 backbone.
Additionally, and contrary to many recent approaches, we do not use a recurrent model such as an LSTM or a GRU to model the interaction between characters, and instead use a purely convolutional model.
The main argument is that recurrent models are slower at inference time, and a small loss in accuracy is affordable if the inference time is reduced.
 Additionally, a few works have started showing that recurrent models are not strictly necessary to accurately model the interaction between characters \cite{DBLP:journals/corr/abs-1709-04303}.

We also evaluated how case sensitive labelling would influence the accuracy of our model.
We found that either ignoring and not ignoring the character case during training led to equivalent results when testing in a case-insensitive manner.
However, training with case-sensitive annotations allowed us to perform case-sensitive inference, which can be useful in some downstream applications of {\systemname}.

The CTC model achieves high accuracy on real-world validation set when trained with synthetic data and fine-tuned
 with manually annotated word crops.
 We also evaluated the total edit distance, \ie, the total number of single character edits in
 the test set needed to correct the predicted transcriptions, as shown in Table \ref{tab:tablerecognitionmetrics}.
 The fully convolutional CTC model (referred to as "CTC, Synthetic $\rightarrow$Human rated"
 in Table \ref{tab:tablerecognitionmetrics}) improved the accuracy by 48.06\% over the CHAR baseline.



We performed evaluation of the system to measure the gap of combining errors in text detection and
text recognition, where in the case of perfect text detection we would get pure text recognition accuracy. We observed that in around 37\% of text detection error cases our text recognition model could still correctly recover words making just one character error. In Table \ref{tab:end2endmetrics}, we show relative drop in recall for overall number of detected words.
For many downstream applications, single-character errors in recognized words are still acceptable
and useful. 

To improve the overall accuracy of the system, we augmented the training dataset for recognition model by
introducing random jittering. The bounding box coordinates of ground-truth might be randomly shifted to model the behavior of
noise from detection model. This resulted in 1.54\% relative improvement in end-to-end performance.
We observed that jittering becomes especially useful when there is less amount of training data available
for certain application use-cases.

\section{Deployment}\label{sec:deployment}
{\systemname} service is deployed within Facebook at scale, offers a cloud API for text extraction from images, and processes a large volume of images uploaded to Facebook and Instagram everyday.
In {\systemname}, the image is resized to 800px in the larger dimension and fed through the detection model which outputs bounding box coordinates around each word. The word patches are then cropped out, resized to a height of 32px while maintaining the aspect ratio, and processed by the recognition model. The inference runtime for the recognition model depends on the number of words detected in the image.

{\systemname} service was deployed incrementally to client applications to anticipate any issues, with such a deployment plan consisting of weekly increase of traffic served initially to a predefined set of internal users, then to public traffic of 1\%, 5\%, 10\%, 25\%, 40\%, 80\% and finally 100\%. We continued to evaluate resource utilization and incrementally add more machines to the processing fleet as deployment of the service continued until 100\%.

The Faster-RCNN detection model was trained using the recently open-sourced Detectron framework~\cite{DetectronOSS} which is built on top of Caffe2 ~\cite{Caffe2Framework}. The text recognition model was trained using PyTorch ~\cite{PyTorchFramework} owing to its flexibility for quick prototyping and sequence modeling scenarios. Both models were deployed to production using Caffe2, with the text recognition model converted from PyTorch to Caffe2 using the intermediate ONNX format ~\cite{onnxFramework}.

\section{Conclusion}\label{sec:conclusion}
In this paper we presented approaches for building robust and efficient models for text detection and recognition, and discussed architectural approaches for building a scalable OCR system {\systemname}. With thorough evaluation, we demonstrated trade-offs between achieving high efficiency in terms of scale and processing time and the accuracy of models. Our system is deployed to production and processes images uploaded to Facebook and Instagram everyday. We have provided practical experience, trade-offs and shared lessons learnt building OCR system at scale.

\section{Acknowledgments}
The authors would like to thank Anmol Kalia, Manohar Paluri, Peizhao Zhang, Sam Tsai, Fei Yang, Vajda Peter, Ross Girshick, Cristian Canton Ferrer, Simon Elmir, Alex Kapranoff, Denis Sheahan, Isaac Nwokocha, Tilak Sharma, Kevin Chen, Nikhil Johri, Shomir Dutt, Cristina Scheau, Yu Cao, Daniel Olmedilla, Brandon Chen, Ainulindale Yeqi Lu, and others who contributed, supported and collaborated with us during the development and deployment of our system.

\balance
\bibliographystyle{abbrv}
\bibliography{bibliography_doc}

\end{document}